\documentclass[11pt,a4paper]{article}
\usepackage[hyperref]{emnlp2018}
\usepackage{times}
\usepackage{latexsym}
\usepackage{graphics} 
\usepackage{graphicx}
\usepackage{url}
\usepackage{float}
\graphicspath{{figures/}}
\aclfinalcopy 

\title{KT-Speech-Crawler: Automatic Dataset Construction for Speech Recognition from YouTube Videos}

\author{Egor Lakomkin \hspace{0.5cm} Sven Magg \hspace{0.5cm} Cornelius Weber\hspace{0.5cm} Stefan Wermter \\
  Department of Informatics, Knowledge Technology\\ University of Hamburg \\ Vogt-Koelln Str. 30, 22527 Hamburg, Germany \\
  {\tt \{lakomkin, magg, weber,  wermter\}@informatik.uni-hamburg.de}
  }

\date{}

\begin{document}
\maketitle
\begin{abstract}
In this paper, we describe KT-Speech-Crawler: an approach for automatic dataset construction for speech recognition by crawling YouTube videos. We outline several filtering and post-processing steps, which extract samples that can be used for training end-to-end neural speech recognition systems. In our experiments, we demonstrate that a single-core version of the crawler can obtain around 150 hours of transcribed speech within a day, containing an estimated 3.5\% word error rate in the transcriptions. Automatically collected samples contain reading and spontaneous speech recorded in various conditions including background noise and music, distant microphone recordings, and a variety of accents and reverberation. When training a deep neural network on speech recognition, we observed around 40\% word error rate reduction on the Wall Street Journal dataset by integrating 200 hours of the collected samples into the training set. The demo\footnote{\url{http://emnlp-demo.lakomkin.me/}} and the crawler code\footnote{\url{https://github.com/EgorLakomkin/KTSpeechCrawler}} are publicly available.
\end{abstract}

\section{Introduction}
\begin{figure}[t]
  \includegraphics[width=\linewidth]{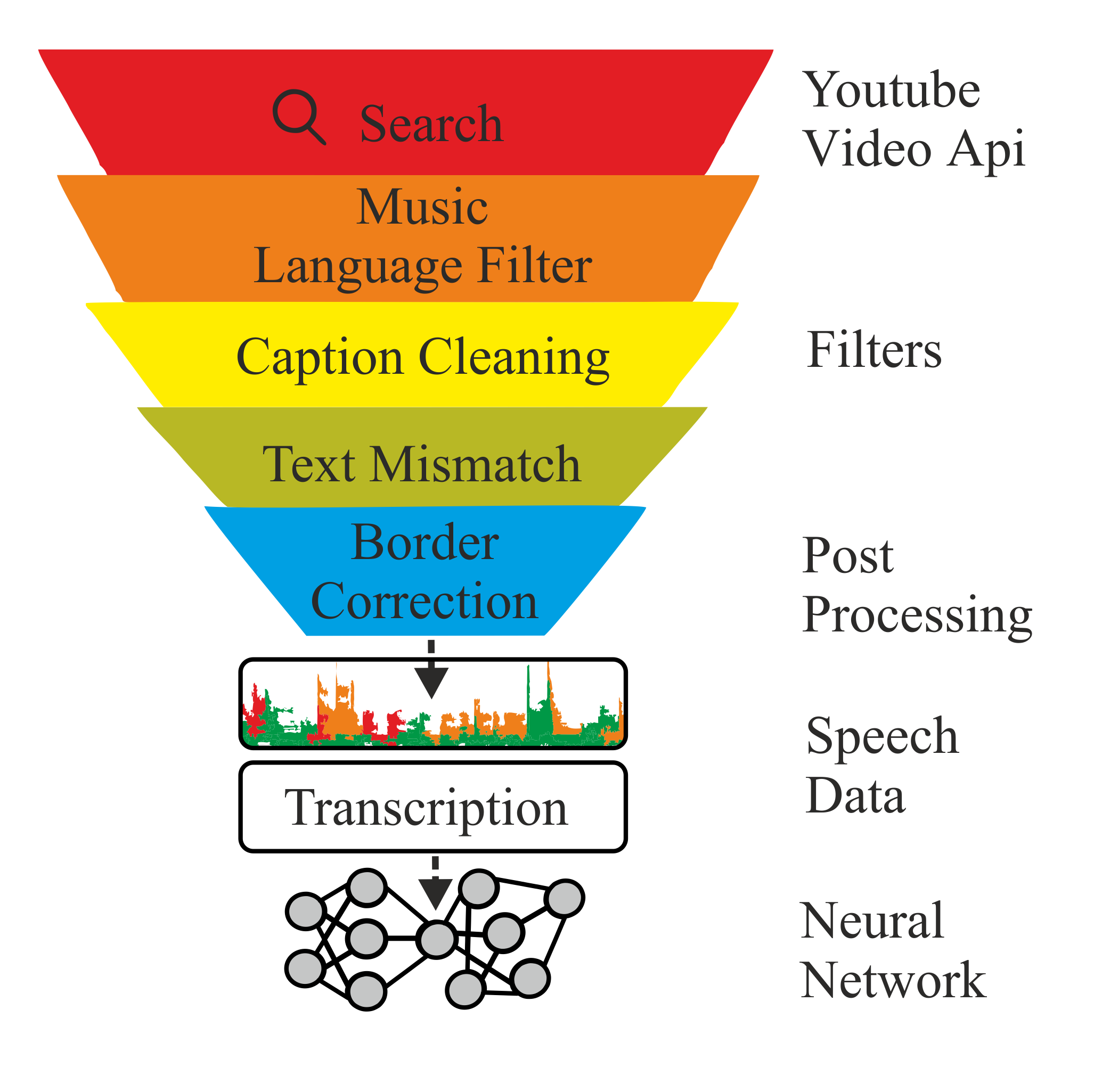}
  \caption{ Architecture of the proposed system crawling YouTube to find videos with closed captions. Several filtering and post-processing steps are applied to select high-quality speech candidates. As a result, pairs of speech and corresponding transcriptions are collected. }
  \label{fig:system_main}
\end{figure}

 \par End-to-end neural networks significantly simplified the development of automatic speech recognition (ASR) systems \cite{Graves2014TowardsNetworks}. Traditionally, ASR systems are based on Gaussian Mixture Models (GMM) or Deep Neural Networks (DNN) for acoustic state representations followed by the Hidden Markov Model (HMM) for sequence-level learning. Though such systems are successful and achieve high performance \cite{Hinton2012DeepGroups}, they require word- or phoneme-level alignments between the acoustic signal and the transcription. As a result, dataset preparation for such hybrid systems is a labor-intensive and error-prone process as the performance of the whole system is sensitive to the quality of the alignment. Also, each component is trained individually, which makes the whole process complex and difficult to maintain. Recently, Connectionist Temporal Classification (CTC) loss \cite{Graves2006ConnectionistClassification}  has been introduced, which allows relaxing the constraint of having alignment between the spoken text and audio by introducing a sequence-level criterion. Also,
 recurrent neural network-based architectures that are state-of-the-art models in machine translation have been applied to speech recognition \cite{Chan2016ListenRecognition}. Consequently, neural networks can be trained end-to-end via backpropagation \cite{Graves2014TowardsNetworks}. CTC maximizes the log-likelihood of the ground truth transcription and thus only the spoken text is required without an explicit alignment, which is easier and cheaper to obtain.
 \par Previous work outlined the importance of having large amounts of annotated data to train deep neural networks. For example, a ten times increase of the training data size from 1,200 hours to 12,000 hours resulted in improving the word error rate from 13.9\% to 8.46\% for clean and from 22.99\% to 13.59\% for noisy speech \cite{Amodei2016DeepMandarin}. Collecting such large datasets is an expensive and labor-intensive process, which requires a significant amount of resources, usually not available for the research community compared to large industrial companies. For example, Baidu's internal speech dataset \cite{Amodei2016DeepMandarin} contains around 10,000 hours of speech, while the largest dataset available for the research community does not exceed 2,000 hours \cite{David2004TheSpeech-to-Text}. We propose to utilize a vast amount of videos available on YouTube with user-provided closed captions as a source to extract speech datasets comparable in size to the ones available in the industry.
 \par Our contribution in this paper is two-fold: 1) we provide a crawler that automatically extracts speech samples with transcriptions from YouTube and filters high-quality samples with several heuristic measures, and 2) we extend the training data of two benchmark datasets with the extracted samples and validate the benefit of the collected data by training a deep neural network on the original and the combined data to measure test performance difference. We also evaluate the amount of noise in transcriptions by manually checking the word error rate of a random subset of the dataset. We hope that our developed tool will foster research of large-scale automatic speech recognition systems\footnote{The code and the Dockerfile are available by this link \url{https://github.com/EgorLakomkin/KTSpeechCrawler}}.

\section{Related work}
 \par Crowdsourcing has been successfully used to construct speech datasets like VoxForge\footnote{\url{http://www.voxforge.org}} or Mozilla's Common Voice\footnote{\url{https://voice.mozilla.org/}}, where users recorded themselves through the provided web-interface, and uploaded samples can be checked by other participants. While such an approach, in theory, can be a viable strategy to acquire a large number of diverse speech samples, it has several drawbacks. The main limitation of this approach is the difficulty of engaging and acquiring users to donate samples to achieve a large and diverse dataset in terms of the number of different speakers, accents, environments and recording conditions. Another approach, which is widely adopted by the research community, is to make use of a vast amount of available multi-modal data which contains transcribed speech. For example, TED talks \cite{Rousseau2014EnhancingTalks} are carefully transcribed and contain around 200 hours of speech from around 1,500 speakers (TED-LIUM v2 dataset). LibriSpeech \cite{Panayotov2015Librispeech:Books} is composed of a large number of audiobooks and is the largest freely available dataset: around 960 hours of English read speech. Google released their Speech Commands dataset\footnote{\url{https://ai.googleblog.com/2017/08/launching-speech-commands-dataset.html}} containing around 65,000 one-second-long utterances.
 \par It has already been demonstrated that YouTube captions can be successfully used as a ground truth spoken text transcription to train large-scale ASR systems \cite{Liao2013LargeTranscription, Lecouteux2012IntegratingCorpora}. Users upload closed captions for various reasons: to make video accessible for people having some degree of hear loss, or to help non-native speakers, or to increase the number of views (YouTube search ranking algorithm indexes closed captions content\footnote{\url{https://www.3playmedia.com/customers/case-studies/discovery-digital-networks/}}). Nevertheless, some videos contain inaccurate or even unrelated to speech captions, for example, advertisements. Several heuristics were proposed to remove low-quality samples: removing captions containing advertisements, language mismatch detection and using forced alignment to detect confident alignment regions between the caption and the audio.  In addition, YouTube has been used previously in multiple ways to automatically collect multi-modal datasets, e.g. emotion recognition datasets by \citet{Barros2018TheDataset} and \citet{Zadeh2016MOSI:Videos}, or opinion mining \cite{Marrese-Taylor2017MiningAttention-RNN}, or video classification (YouTube-8M\footnote{\url{https://research.google.com/youtube8m/}}, or human action recognition \cite{Kay2017TheDataset}).
  \par In this work, we combine several known heuristics and propose some additional ones to select high quality samples in an automatic way. We integrate it into an easy to use tool \textit{KT-Speech-Crawler}, which can continuously scan new videos uploaded to YouTube and update the speech database. To our knowledge, this is the first open-source tool available for automatic speech dataset construction.
 
\section{Crawler}
In this section, we describe the sample selection strategy, followed by several filtering and post-processing heuristics to locate high-quality samples and discard noisy ones from YouTube.

\subsection{Candidate selection}

Firstly, we download candidate videos with English closed captions, which are usually uploaded by the channel owner. To reach as many videos as possible we use the YouTube Search API, where one of the top 100 most common English words is used as a search keyword to match the video title (for example, \textit{the, but, have, not, and, ...}). Such frequent keywords allow us to match many videos, even though, as a side effect, non-English videos with closed captions in English might be captured. The YouTube Search API allows to download the 600 most recent videos for each keyword, and since many videos are constantly being uploaded to YouTube it is possible to continuously collect speech samples. Also, we memorize YouTube channels containing samples that passed all the filtering steps (see section 3.2) and use other videos from this channel. This leads to many diverse candidates coming from TV shows and TV series, video blogs, news, and live recordings.

\subsection{Filtering steps}

We perform several filtering steps to select suitable candidates:
\begin{itemize}
\item we discard a caption if it overlaps with another caption, which sometimes happens due to incorrectly closed caption auto syncing,
\item we filter out captions that indicate that there is music content in this sample and captions containing non-ASCII characters or URLs,
\item we remove text chunks which do not correspond to the actual spoken text, like the information of the speaker name (\textit{Speaker 1: ...}), annotations (\textit{[laughs], *laughs*, (laughs)}), and punctuation,
\item we spell out numbers which are within the range from 1 to 100 as they have non-ambiguous pronunciation (in contrast, for example, \textit{1,500} can be uttered as \textit{fifteen hundred} or \textit{one thousand and five hundred}),
\item we discard captions if they contain any character that is not an English letter, apostrophe or a white space,
\item we filter segments which have less than one second duration or more than ten seconds,
\item in addition, we select randomly three phrases from the video and measure the Levenshtein similarity between the provided closed caption and the transcription generated by the Google ASR API. If the similarity is below a 70\% threshold, we discard all the samples in this video. This step allows filtering videos which have English subtitles for non-English spoken text or videos with a bad alignment. Also, this filter removes videos with completely misaligned captions.
\end{itemize}

\subsection{Post-processing steps}
During our experiments on evaluating the quality of the extracted samples, we spotted that one of the major problems is imprecise alignments between caption and audio. For example, the first or the last word can be omitted on the recording due to incorrect caption timings. One possible way to reduce the number of samples with misaligned borders is to group together nearby captions if they are at a distance of less than one second. We stop grouping adjacent utterances if the overall length exceeds ten seconds. In addition, we perform a forced alignment\footnote{\url{https://github.com/lowerquality/gentle}} between the caption and the corresponding audio using Kaldi \cite{Povey2011TheToolkit} and if the first or the last word is not successfully mapped, we try to extend the caption boundaries (up to 500 milliseconds) until the border word becomes mapped. If we cannot align the border word, we keep the caption boundaries unchanged.

\begin{figure}
  \includegraphics[width=\linewidth]{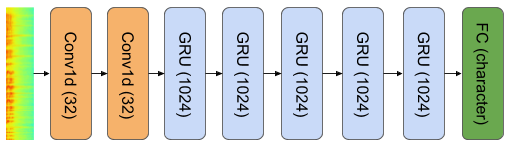}
  \caption{Architecture of the ASR model used in this work, following the DeepSpeech 2 architecture.  }
  \label{fig:asrdeepspeech}
\end{figure}

\section{Experiments and analysis}
To evaluate the usefulness of the collected samples we conducted three types of experiments. We trained the deep neural network-based model on different training datasets:
\begin{itemize}
\item on the original training data,
\item on the mix of the original and with the crawled samples,
\item only on the crawled samples.
\end{itemize}
For benchmarking, we selected two well-known datasets for training ASR systems: The Wall Street Journal and TED-LIUM v2. In all experiments, we kept the same size and architecture of the neural model and its hyperparameters. In this section, we outline the details of benchmark data used in our experiments, neural model architecture and the evaluation protocol and metrics, followed by the evaluation results and comparisons. 

\subsection{ASR model}

Our ASR model (see Figure \ref{fig:asrdeepspeech}) is a combination of convolutional and recurrent layers inspired by the DeepSpeech 2 \cite{Amodei2016DeepMandarin} architecture. Our model contains two 2D convolutional layers for feature extraction from power FFT spectrograms. Power spectrograms are extracted using a Hamming window of 20ms width and 10ms stride, resulting in 161 features for each speech frame. Convolutional layers are followed by five recurrent bi-directional Gated Recurrent Units \cite{Chung2014EmpiricalModeling} layers with a size of 1,024 followed by a softmax layer on top, predicting the character distribution for each speech frame. Overall, our model has around 61 million parameters. Connectionist Temporal Classification (CTC) loss \cite{Graves2006ConnectionistClassification} is used as a loss criterion to measure how good the alignment produced by the network is compared to the ground truth transcription. \par The Stochastic Gradient Descent optimizer is used in all experiments with a learning rate of 0.0003, clipping the norm of the gradient at the level of 400 with a batch size of 32. During the training, we apply learning rate annealing with a factor of 1.1. We apply the SortaGrad algorithm \cite{Amodei2016DeepMandarin} during the first epoch by sorting utterances by their duration \cite{Hannun2014DeepRecognition}. We select the model with the best word error rate measured on the validation set to prevent model overfitting.

\begin{table}[t]
\centering
\caption{Evaluation results. We evaluated the effect of adding samples extracted from YouTube by our tool on two benchmarking datasets: WSJ and TED-LIUM v2. We trained the deep neural network on the original training data, then combined the data with YouTube samples (\textit{WSJ+YouTube}, for example), and, finally, only on the YouTube samples. We report word and character error rate. }
\label{results}
\begin{tabular}{llll}
\multicolumn{1}{l|}{Train} & \multicolumn{1}{l|}{Test} & \multicolumn{1}{l|}{WER} & CER \\ \hline
WSJ                                   & WSJ                               & 27.4\%                                 & 7.2\%                 \\
WSJ + YouTube (200h)                  & WSJ                               & \textbf{15,8\%}                                 & \textbf{4.2}\%                 \\
YouTube (200h)                        & WSJ                               & 31.5\%                                 & 8.3\%                 \\ \hline
TED                                   & TED                               & 32.6\%                                 & 10.4\%                 \\
TED + YouTube (300h)                  & TED                               & \textbf{28.1\%}                                 & \textbf{8.2\%}                 \\
YouTube (300h)                        & TED                               & 36.6\%                                 & 10.6\%                
\end{tabular}
\end{table}

\subsection{Data and evaluation measure}

\subsubsection{WSJ}
\par The Wall Street Journal (WSJ) dataset is a well-known dataset for evaluating ASR systems, containing utterances of read speech coming from the news domain. The WSJ training set (\textit{’train-si284’}) consists of 81 hours  containing 37,318 sentences from 284 speakers (142 male and 142 female). We used the ’dev93’ development set for validation and report the word error rate on the ’eval92’ test set.

\subsubsection{TED talks}
\par We also evaluated our approach on the TED-LIUM v2 dataset, which contains around 200 hours of transcribed TED\footnote{\url{https://www.ted.com/}} talks of 1,495 speakers. In contrast to the WSJ dataset, it contains spontaneous speech rather than read speech.

\subsection{Results}
\par We summarize our results in Table \ref{results}. Note that we did not use a language model for decoding in our experiments but used greedy decoding, where the most probable character at each timestep was emitted. It is well known that decoding with the language model and beam search significantly improves the performance on the test set of character-based end-to-end models \cite{Hannun2014First-PassDNNs}, but as our goal was to demonstrate the impact of adding extracted samples within the same neural model and test set, we left it out. We observed that adding samples from YouTube positively contributed to the overall performance in both metrics: word (WER) and character error rates (CER). For example, the word error rate improved from 34.2\% to 15.8\% on the WSJ test set by adding 200 hours of samples (108,617 utterances) to the WSJ training set. Similar results can be observed on the TED talks dataset: WER and CER improved from 32.6\% to 28.1\% and 10.4\% to 8.2\% by adding 300 hours of YouTube samples. To be sure that none of the TED videos appeared in the YouTube set, which could lead to overestimation of the performance, we excluded videos that contain a \textit{TED} token in the title or in the description.  Interestingly, if only YouTube samples were used as the training set, we observed CER values of 8.3\% and 10.6\% for the WSJ and the TED datasets, respectively (compared to 7.2\% and 10.4\% using original training data), indicating that having a domain-specific training set plays an important role and there is a room for improvement in designing better filtering and post-processing steps.

\subsection{Transcriptions quality}
\begin{figure}
  \includegraphics[width=\linewidth]{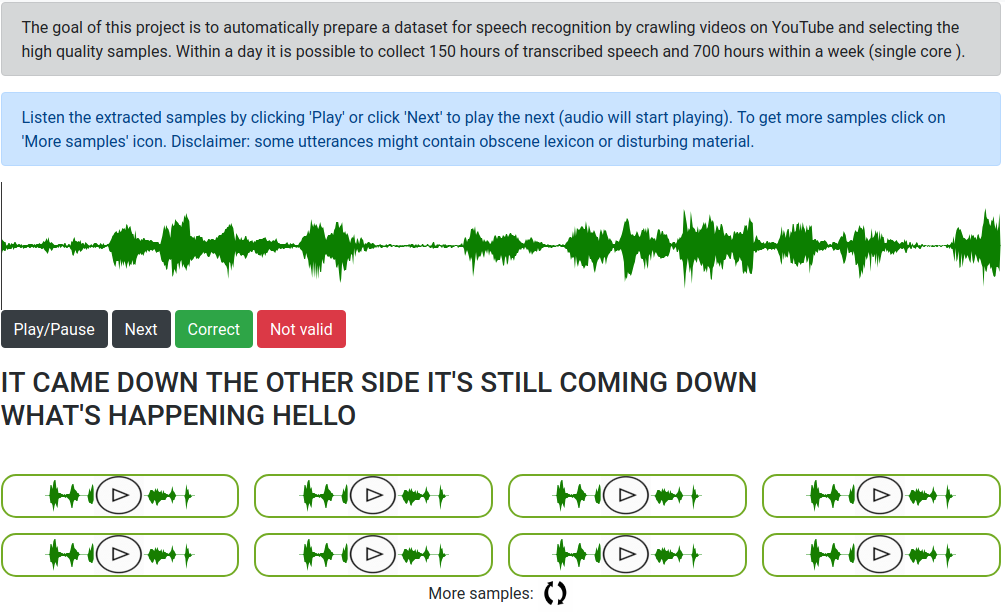}
  \caption{ A screenshot of the web-based demo to browse the collected samples, presenting the extracted utterance and the corresponding transcription.  }
  \label{fig:demo}
\end{figure}
We manually investigated samples by using developed a web-based demo, see Fig. \ref{fig:demo} and analyzed the quality of the collected samples and their transcriptions. Our developed web-service presents random eight utterances and their corresponding transcriptions to the user and allows to load more samples if necessary. 
We also integrated a simple functionality to validate the extracted samples: a user can confirm that the caption is correct or if not enter the right transcription. 
\begin{equation}
\label{eq.wer}
WER = \frac{S+D+I}{S+D+C}
\end{equation}
We computed the word error rate using equation \ref{eq.wer}, where \textit{S, I, D, C} is number of substitutions, insertions, deletions and correct words, respectively. We estimated 3.5\% word error rate on the small randomly selected subset of 600 samples. The most common type of error was missing or wrongly added one or two words at the beginning or at the end of the utterance.

\section{Conclusions and future work}

\par In this work, we presented an open-source system that automatically constructs datasets for training end-to-end neural speech recognition systems. We demonstrated the usefulness of the collected samples on the WSJ and TED datasets. We provide the code for the crawler and metadata and a script to easily construct a dataset of 500 hours.
\par Future work includes extending the script to support other languages. A more sophisticated approach to identify wrongly added or missing words in transcriptions could also be used by using attention-based neural networks like pointer networks. We are also aware that some collected samples may contain automatically generated utterances with Text-To-Speech software, which may require performing speaker recognition to balance the dataset. Furthermore, domain-specific speech datasets can be collected by selecting samples after analyzing captions and video metadata (for example, in the financial domain). In addition, samples with several people talking at the same time and noisy samples with low signal-to-noise ratio need to be filtered, which could be implemented as neural network-based modules.
\par We believe that having large, free and high-quality speech datasets available to the research community will foster the development of new architectures and applications for speech understanding, and we hope that our presented tool will contribute to that.

\section*{Acknowledgments}
This project has received funding from the European Union's Horizon 2020 research and innovation programme under the Marie Sklodowska-Curie grant agreement No 642667 (SECURE) and partial support from the German Research Foundation DFG under project CML (TRR 169).

\bibliographystyle{acl_natbib_nourl}
\bibliography{emnlp2018}

\begin{thebibliography}{18}
\expandafter\ifx\csname natexlab\endcsname\relax\def\natexlab#1{#1}\fi

\bibitem[{Amodei et~al.(2016)Amodei, Anubhai, Battenberg, Case, Casper,
  Catanzaro, Chen, Chrzanowski, Coates, and {et al}}]{Amodei2016DeepMandarin}
Dario Amodei, Rishita Anubhai, Eric Battenberg, Carl Case, Jared Casper, Bryan
  Catanzaro, Jingdong Chen, Mike Chrzanowski, Adam Coates, and {et al}. 2016.
\newblock {Deep Speech 2: End-to-End Speech Recognition in English and
  Mandarin}.
\newblock \emph{Proceedings of The 33rd International Conference on Machine
  Learning}, 48:173--182.

\bibitem[{Barros et~al.(2018)Barros, Churamani, Lakomkin, Siqueira, Sutherland,
  and Wermter}]{Barros2018TheDataset}
Pablo Barros, Nikhil Churamani, Egor Lakomkin, Henrique Siqueira, Alexander
  Sutherland, and Stefan Wermter. 2018.
\newblock {The OMG-Emotion Behavior Dataset}.
\newblock \emph{To appear in International Joint Conference on Neural
  Networks}.

\bibitem[{Chan et~al.(2016)Chan, Jaitly, Le, and
  Vinyals}]{Chan2016ListenRecognition}
William Chan, Navdeep Jaitly, Quoc Le, and Oriol Vinyals. 2016.
\newblock {Listen, attend and spell: A neural network for large vocabulary
  conversational speech recognition}.
\newblock In \emph{2016 IEEE International Conference on Acoustics, Speech and
  Signal Processing (ICASSP)}, pages 4960--4964. IEEE.

\bibitem[{Chung et~al.(2014)Chung, Gulcehre, Cho, and
  Bengio}]{Chung2014EmpiricalModeling}
Junyoung Chung, Caglar Gulcehre, KyungHyun Cho, and Yoshua Bengio. 2014.
\newblock {Empirical Evaluation of Gated Recurrent Neural Networks on Sequence
  Modeling}.
\newblock \emph{Deep Learning and Representation Learning Workshop}.

\bibitem[{David et~al.(2004)David, David, Miller, and
  Walker}]{David2004TheSpeech-to-Text}
Christopher~Cieri David, Christopher~Cieri David, David Miller, and Kevin
  Walker. 2004.
\newblock {The Fisher Corpus: a Resource for the Next Generations of
  Speech-to-Text}.
\newblock \emph{In Proceedings 4th International Conference On Language
  Resources and Evaluation}, pages 69--71.

\bibitem[{Graves et~al.(2006)Graves, Fern{\'{a}}ndez, Gomez, and
  Schmidhuber}]{Graves2006ConnectionistClassification}
Alex Graves, Santiago Fern{\'{a}}ndez, Faustino Gomez, and Jürgen Schmidhuber.
  2006.
\newblock {Connectionist temporal classification}.
\newblock In \emph{Proceedings of the 23rd international conference on Machine
  learning - ICML '06}, pages 369--376, New York, New York, USA. ACM Press.

\bibitem[{Graves and Jaitly(2014)}]{Graves2014TowardsNetworks}
Alex Graves and Navdeep Jaitly. 2014.
\newblock {Towards end-to-end speech recognition with recurrent neural
  networks}.
\newblock \emph{Proceedings of the 31st International Conference on
  International Conference on Machine Learning - Volume 32}, pages II--1764.

\bibitem[{Hannun et~al.(2014{\natexlab{a}})Hannun, Case, Casper, Catanzaro,
  Diamos, Elsen, Prenger, Satheesh, Sengupta, Coates, and
  Ng}]{Hannun2014DeepRecognition}
Awni Hannun, Carl Case, Jared Casper, Bryan Catanzaro, Greg Diamos, Erich
  Elsen, Ryan Prenger, Sanjeev Satheesh, Shubho Sengupta, Adam Coates, and
  Andrew~Y Ng. 2014{\natexlab{a}}.
\newblock {Deep Speech: Scaling up end-to-end speech recognition}.
\newblock \emph{CoRR}, abs/1412.5567.

\bibitem[{Hannun et~al.(2014{\natexlab{b}})Hannun, Maas, Jurafsky, and
  Ng}]{Hannun2014First-PassDNNs}
Awni~Y. Hannun, Andrew~L. Maas, Daniel Jurafsky, and Andrew~Y. Ng.
  2014{\natexlab{b}}.
\newblock {First-Pass Large Vocabulary Continuous Speech Recognition using
  Bi-Directional Recurrent DNNs}.
\newblock \emph{CoRR}, abs/1408.2873.

\bibitem[{Hinton et~al.(2012)Hinton, Deng, Yu, Dahl, Mohamed, Jaitly, Senior,
  Vanhoucke, Nguyen, Sainath, and Kingsbury}]{Hinton2012DeepGroups}
Geoffrey Hinton, Li~Deng, Dong Yu, George Dahl, Abdel-rahman Mohamed, Navdeep
  Jaitly, Andrew Senior, Vincent Vanhoucke, Patrick Nguyen, Tara Sainath, and
  Brian Kingsbury. 2012.
\newblock {Deep Neural Networks for Acoustic Modeling in Speech Recognition:
  The Shared Views of Four Research Groups}.
\newblock \emph{IEEE Signal Processing Magazine}, 29(6):82--97.

\bibitem[{Kay et~al.(2017)Kay, Carreira, Simonyan, Zhang, Hillier,
  Vijayanarasimhan, Viola, Green, Back, Natsev, Suleyman, and
  Zisserman}]{Kay2017TheDataset}
Will Kay, Joao Carreira, Karen Simonyan, Brian Zhang, Chloe Hillier, Sudheendra
  Vijayanarasimhan, Fabio Viola, Tim Green, Trevor Back, Paul Natsev, Mustafa
  Suleyman, and Andrew Zisserman. 2017.
\newblock {The Kinetics Human Action Video Dataset}.

\bibitem[{Lecouteux et~al.(2012)Lecouteux, Linar{\`{e}}s, and
  Oger}]{Lecouteux2012IntegratingCorpora}
Benjamin Lecouteux, Georges Linar{\`{e}}s, and Stanislas Oger. 2012.
\newblock {Integrating imperfect transcripts into speech recognition systems
  for building high-quality corpora}.
\newblock \emph{Computer Speech {\&} Language}, 26(2):67--89.

\bibitem[{Liao et~al.(2013)Liao, McDermott, and
  Senior}]{Liao2013LargeTranscription}
Hank Liao, Erik McDermott, and Andrew Senior. 2013.
\newblock {Large scale deep neural network acoustic modeling with
  semi-supervised training data for YouTube video transcription}.
\newblock In \emph{2013 IEEE Workshop on Automatic Speech Recognition and
  Understanding}, pages 368--373.

\bibitem[{Marrese-Taylor et~al.(2017)Marrese-Taylor, Balazs, and
  Matsuo}]{Marrese-Taylor2017MiningAttention-RNN}
Edison Marrese-Taylor, Jorge Balazs, and Yutaka Matsuo. 2017.
\newblock {Mining fine-grained opinions on closed captions of YouTube videos
  with an attention-RNN}.
\newblock In \emph{Proceedings of the 8th Workshop on Computational Approaches
  to Subjectivity, Sentiment and Social Media Analysis}, pages 102--111,
  Stroudsburg, PA, USA. Association for Computational Linguistics.

\bibitem[{Panayotov et~al.(2015)Panayotov, Chen, Povey, and
  Khudanpur}]{Panayotov2015Librispeech:Books}
Vassil Panayotov, Guoguo Chen, Daniel Povey, and Sanjeev Khudanpur. 2015.
\newblock {Librispeech: An ASR corpus based on public domain audio books}.
\newblock In \emph{2015 IEEE International Conference on Acoustics, Speech and
  Signal Processing (ICASSP)}, pages 5206--5210. IEEE.

\bibitem[{Povey et~al.(2011)Povey, Povey, Ghoshal, Boulianne, Goel, Hannemann,
  Qian, Schwarz, and Stemmer}]{Povey2011TheToolkit}
Daniel Povey, Daniel Povey, Arnab Ghoshal, Gilles Boulianne, Nagendra Goel,
  Mirko Hannemann, Yanmin Qian, Petr Schwarz, and Georg Stemmer. 2011.
\newblock {The kaldi speech recognition toolkit}.
\newblock \emph{In IEEE Automatic Speech Recognition and Understanding
  Workshop}.

\bibitem[{Rousseau et~al.(2014)Rousseau, Rousseau, Del{\'{e}}glise, and
  Est{\`{e}}ve}]{Rousseau2014EnhancingTalks}
Anthony Rousseau, Anthony Rousseau, Paul Del{\'{e}}glise, and Yannick
  Est{\`{e}}ve. 2014.
\newblock {Enhancing the TED-LIUM corpus with selected data for language
  modeling and more TED talks}.
\newblock \emph{In Proceedings 9th International Conference On Language
  Resources and Evaluation}, pages 26--31.

\bibitem[{Zadeh et~al.(2016)Zadeh, Zellers, Pincus, and
  Morency}]{Zadeh2016MOSI:Videos}
Amir Zadeh, Rowan Zellers, Eli Pincus, and Louis-Philippe Morency. 2016.
\newblock {MOSI: Multimodal Corpus of Sentiment Intensity and Subjectivity
  Analysis in Online Opinion Videos}.
\newblock \emph{IEEE Intelligent Systems}, 31.6:82--88.

\end{thebibliography}

\end{document}